\documentclass[sigconf,nonacm]{acmart}
\usepackage{graphicx}
\usepackage{color}
\usepackage{booktabs}
\usepackage{amsmath}
\usepackage{hyperref}
\usepackage{makecell}
\usepackage{multirow}
\usepackage{algorithm}
\usepackage{algorithmic}
\usepackage{subcaption}
\usepackage{mathrsfs}
\usepackage{libertine}
\DeclareMathOperator*{\argmin}{argmin}
\DeclareMathOperator*{\argmax}{argmax}
\usepackage{threeparttable}
\usepackage{float}

\AtBeginDocument{%
  \providecommand\BibTeX{{%
    \normalfont B\kern-0.5em{\scshape i\kern-0.25em b}\kern-0.8em\TeX}}}

\begin{document}

\title{Customizable and Jointly Optimized Route Planning: A Deep Architecture Enabling Differentiable Shortest-Path Search}

\author{Rui Zhao}
\affiliation{%
	\institution{Alibaba Group}
	\city{Beijing}
	\country{China}
}
\email{melody.zr@alibaba-inc.com}

\author{Chao Chen}
\affiliation{%
	\institution{Alibaba Group}
	\city{Beijing}
	\country{China}
}
\email{cc201598@alibaba-inc.com}

\author{Longfei Xu}
\affiliation{%
	\institution{Alibaba Group}
	\city{Beijing}
	\country{China}
}
\email{longfei.xl@alibaba-inc.com}

\author{Chenguang Ji}
\authornote{Corresponding author.}
\affiliation{%
	\institution{Alibaba Group}
	\city{Beijing}
	\country{China}
}
\email{chenguang.jcg@alibaba-inc.com}

\author{Hengbin Cui}
\affiliation{%
	\institution{Alibaba Group}
	\city{Hangzhou}
	\country{China}
}
\email{alexcui.chb@alibaba-inc.com}

\author{Kaikui Liu}
\affiliation{%
	\institution{Alibaba Group}
	\city{Beijing}
	\country{China}
}
\email{damon@alibaba-inc.com}

\author{Xiaolong Li}
\affiliation{%
	\institution{Alibaba Group}
	\city{Hangzhou}
	\country{China}
}
\email{xl.li@alibaba-inc.com}


\begin{abstract}
With the widespread use of online navigation and ride-hailing services, achieving optimal route planning for diverse user preferences has recently attracted increasing attention. Classic graph algorithms for pathfinding use heuristic cost functions to define edge weight, thus providing no optimality guarantee of route quality. Prior data-driven approaches equating ground truth of the optimal route with user trajectory, which is however moderately influenced by the navigation service, suffers from the feedback loop problem. To address these issues, we propose a deep architecture that is able to jointly optimize cost functions and route-ranking model towards any route preference. First, we run a multi-objective Dijkstra algorithm offline to collect the set of Pareto optimal routes, deeming it as the complete candidate set. Exploiting the property of such a set, we design a neural network structure that emulates shortest-path search and route ranking in an end-to-end differentiable manner. Second, we define route preference as a task of constrained optimization of route attributes, and propose a novel loss function that optimizes a single-objective variable, with other variables strictly under constraints. We conduct extensive experiments on real-world datasets. The results show that our architecture significantly outperforms state-of-the-art methods in route quality and customizability.
\end{abstract}


\maketitle

\section{Introduction}
Route planning aims to find the optimal path between two locations. It has widespread applications in modern industry and plays an especially crucial role in high-tech fields, such as real-time GPS navigation in map apps, on-demand ride-hailing service, intelligent logistics transportation systems, etc. Performance of route planning delivers significant social and economic impacts in daily life.

\begin{figure}[h]
	\centering
	\begin{minipage}[t]{1\linewidth}
		\begin{subfigure}[t]{0.48\textwidth}
			\centering
			\includegraphics[width=\linewidth]{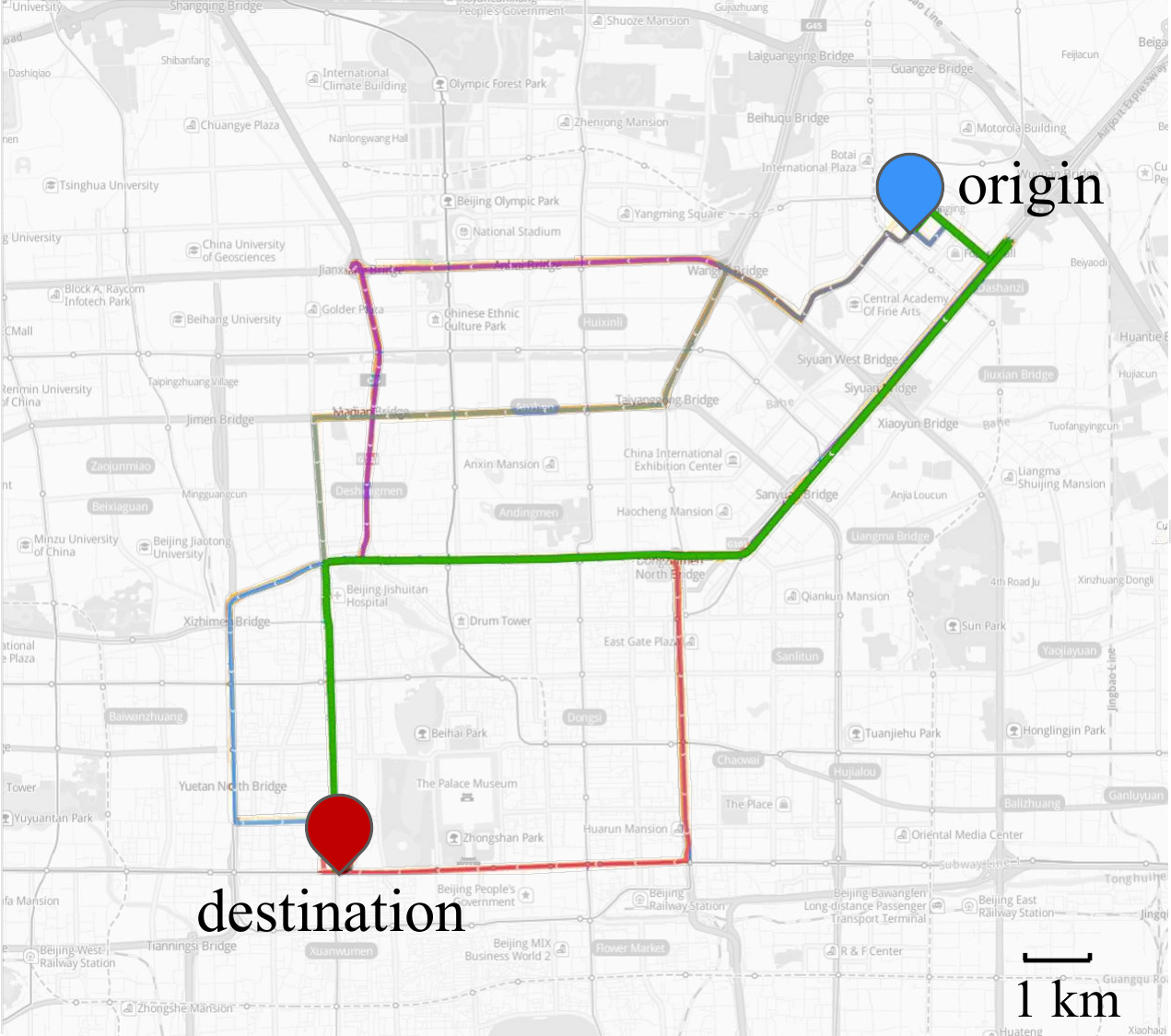} 
			\label{fig:pareto_recall_case_a}
		\end{subfigure}
		\hfill
		\begin{subfigure}[t]{0.48\textwidth}
			\centering
			\includegraphics[width=\linewidth]{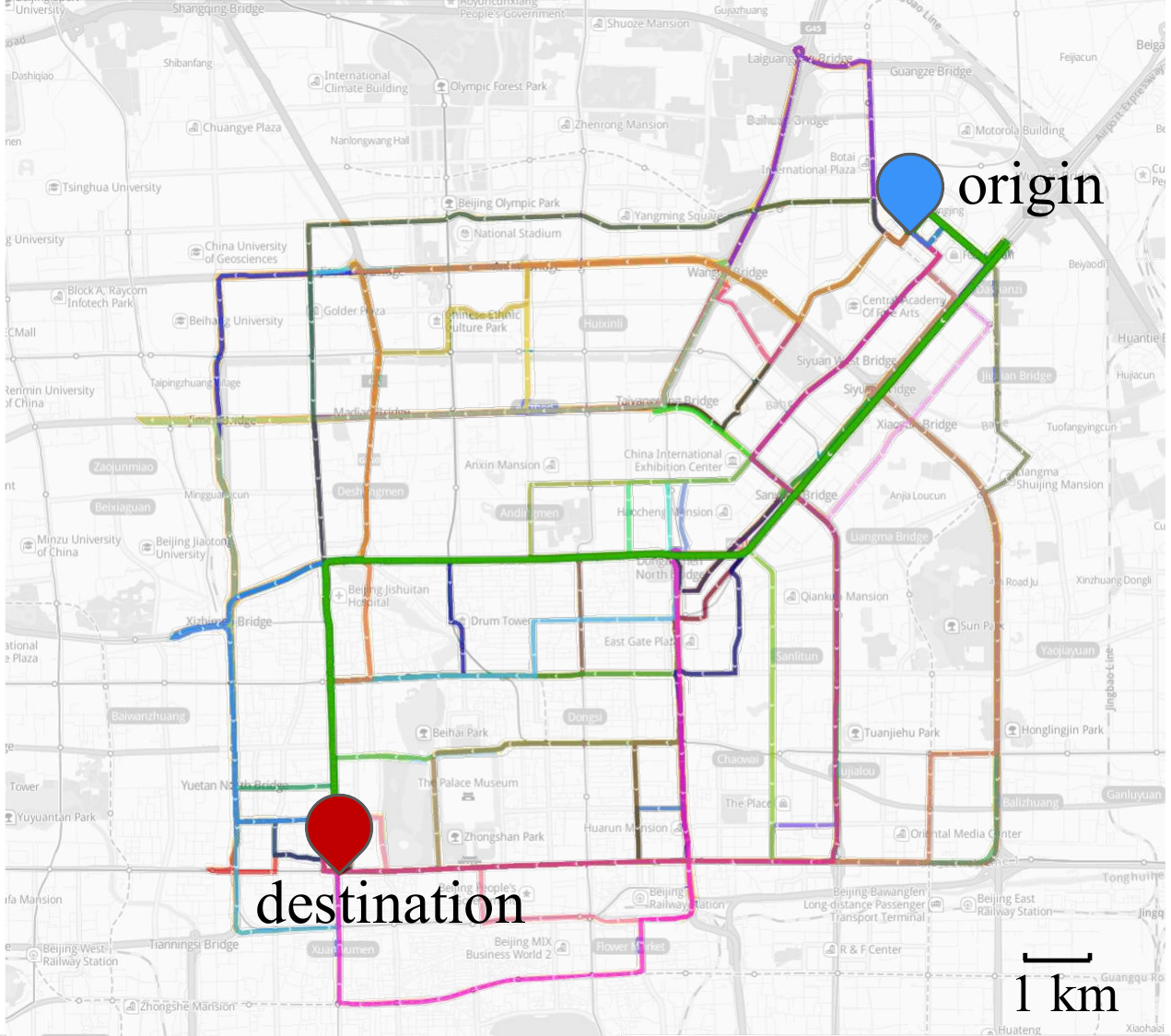} 
			\label{fig:pareto_recall_case_d}
		\end{subfigure}
	\end{minipage} 
	\caption{Comparison of routes recalled with single-objective Dijkstra (left) and multi-objective Dijkstra (right).}
	\label{fig:route_comparison}
\end{figure}

Challenges in real-world route planning lie in two aspects. First, given an origin-destination pair (OD pair), there are infinite possible routes in-between, and the ones recalled hinge on a set of cost functions (a mapping from an edge in a graph to a real-valued weight), which are usually predetermined heuristically \cite{bast2016route} and thus have no optimality guarantee. One way to get around this issue is to use multi-objective routing \cite{martins1984multicriteria}. In multi-objective optimization \cite{marler2004survey}, a solution is called Pareto optimal if none of the objective functions can be improved in value without degrading some of the other objective values. Therefore, a Pareto-optimal route set can be deemed as a complete candidate set, as the full commodity set in e-commerce, in that it contains the optimal solution for any route preference built upon these objectives. The idea is illustrated in Figure \ref{fig:route_comparison}. However, multi-objective routing is not polynomially solvable and thus cannot be applied in real-time services. Second, provided that users of navigation apps tend to follow the recommended (top-ranked) route, user trajectories are often biased with respect to their true preference. As a consequence, prior formalism \cite{wang2019empowering} that uses user trajectory as the supervising signal suffers from the feedback loop problem \cite{sinha2017deconvolving} and neglects to construct a rigorous definition for route preference, and is thus unable to customize route planning as needed.

To solve the first issue, we propose an end-to-end learning framework consisting of a differentiable shortest-path search module and a following route-ranking module. In real-world route planning, though many possible routes exist between an OD pair, only a few of them are practical to users. Inspired by this observation, we design a differentiable shortest-path search structure to learn several cost functions, whose shortest paths constitute a condensed version of the Pareto-optimal set that suffices to meet user demands. Specifically, we first run the multi-objective Dijkstra algorithm offline to collect the Pareto-optimal set for each routing request (OD pair). Adopting linear combination of these objectives as the learnable cost function (scoring metrics), we then transfer the shortest-path problem into an equivalent ranking problem: the lowest-scoring route in the Pareto-optimal set is theoretically guaranteed to be the shortest route under the scoring metrics. As such, a soft indicator vector of the recalled routes is constructed using the learnable weights of the cost functions. At last, the ranking module takes as input the indicator-multiplied route features and produces the ranking score.

To solve the second issue, we propose a formal definition for route preference, which is immune to the position bias. In daily experience, users' route preference could be usually phrased in a format describing a constrained optimization task. For example, “avoid congestion” may mean “prefer a route with the least congestion as long as its travel time and distance are no more than 1.2 times the minimum values”. This motivates us to equate each route preference as a customized loss function, with a single-objective function and a few constraints. Moreover, since route attribute, such as travel time, is innately independent of the ranking strategy, there exists an unbiased estimator trainable on user trajectories for each attribute. We build these estimators and apply them to construct the objective function and constraints, naturally avoiding the feedback loop problem.

To further suit the above-mentioned constraints for practical application, we propose a novel batch sampling method. In industrial navigation services, routing requests with disparate parameters correspond to different travel scenarios. For instance, a request with 10 km OD-distance issued at peak hour is likely a daily commute, while another request with 5 km distance at night might be a casual shopping trip. Therefore, the constraints must be enforced locally across the parametric space of requests to avoid severely damaging the routing performance on a certain portion of requests. To serve this purpose, we use a two-stage batch sampling method in each iteration: first, randomly sample an anchor point in the parametric space, and then randomly sample a batch of requests within a window centered around the anchor as the training data.

In this paper, we propose an end-to-end machine learning solution for route planning that overcomes the drawbacks of the existing methods. We build a module to find the shortest path in a differentiable manner. In this module, the Pareto-optimal set are collected offline via a multi-objective Dijkstra algorithm. Using linear combinations of these objectives as the learnable cost functions, we transform shortest-path search into an equivalent ranking problem within the Pareto-optimal set. We design a deep neural network structure to emulate such a process and produce a soft-indicator vector of the associated shortest routes. The following rank module applies the same indicator technique to further extract the top-ranked route and feed its attributes into the loss function. Particularly, the loss function is customized in line with the route preference, with an objective function and a few constraints. Variables in the loss function, which represent route attributes, are predicted with estimators inferred from historical user trajectories. This architecture offers several advantages. First, the learned cost functions are guaranteed to be optimal towards the route preference. Second, it is computationally fast enough to be applied in real-time services. Third, it gets rid of the problem of position bias, and can be customized to adapt to any route preference. The results of city-scale online A/B testing and offline evaluation demonstrate that the proposed solution outperforms state-of-the-art methods in both route quality and customizability.

The contribution of this paper can be summarized as follows:
\begin{itemize}
	\item We propose a novel end-to-end learning framework for route planning, which is able to jointly optimize the cost functions and route-ranking model.
	\item We define route preference as a constrained optimization task, embodied in a customizable loss function.
	\item We propose a novel batch sampling method to enforce local constraints in a machine learning task.
	\item We evaluate our proposed solution with city-scale online A/B testing and offline data from Amap.\footnote{Amap is the top-tier LBS-service provider in China, and served more than 150 million users on the National Day of the People’s Republic of China in 2020, according to a third-party report https://www.questmobile.com.cn/en.} The results show that the proposed framework significantly outperforms state-of-the-art methods in both route quality and customizability.
\end{itemize}

The rest of the paper is organized as follows: Section \ref{related_work} reviews the related works. Section \ref{preliminaries} outlines the preliminary concepts and formulates the joint optimization problem of route planning. Section \ref{methodology} details the structure of the proposed learning framework. Section \ref{experiment} describes the results of the experiment. Finally, Section \ref{conclusion} concludes the paper.

\section{RELATED WORK\label{related_work}}
Our work is related to the following research directions:

{\bfseries Graph Algorithms for Optimal Path Finding.} Route planning has been studied for decades. In the early days, the core issue was to speed up the shortest-path finding for real-world application \cite{bast2016route}. Goal-directed techniques, such as A star \cite{hart1968formal}, ALT \cite{goldberg2005computing}, etc. tactically guide the search toward the target and obtain a considerable acceleration. Contraction hierarchy \cite{geisberger2012exact} and customizable route planning (CRP) \cite{delling2017customizable}—the preprocessing-based methods—compute reusable shortcuts in advance and further achieve milliseconds of query response time on continental-scale road networks. These approaches, though compatible with any common metrics, such as distance, discuss little about how to optimize the cost function. To overcome this limitation, instead of a single optimal route, multi-objective routing \cite{martins1984multicriteria, delling2015round} computes the Pareto-optimal set that offers the best achievable route diversity with respect to the objectives. Multi-objective routing has been successfully deployed in public-transit routing \cite{delling2015round}. Nonetheless, this routing approach is infeasible for larger road networks, due to its exponential time complexity. 

{\bfseries Personalized Route Planning.} Assuming historical trajectories fully reveal users' route preferences, prior works either develop algorithms to directly search the most frequently travelled patterns \cite{luo2013finding}, or utilize machine learning techniques to capture users' transition probability between locations \cite{wang2019empowering}. However, this basic assumption contradicts on-the-ground realities: (1) over 95\% of online-navigation users follow the recommended (top-ranked) route, and their behaviors are thereby moderately influenced by the routing strategy of the map app; (2) users tend to drive the familiar route, which is not necessarily in their best interest.

{\bfseries Recommender System.} Recommender systems have long been deployed in e-commence \cite{schafer2001commerce}, video-sharing platforms \cite{zhao2019recommending}, etc. With finite candidate items, prior approaches primarily focus on optimizing the ranking model. To reconcile conflicting objectives, such as clicks, watches, likes and dismissals on video-sharing platforms, \citet{zhao2019recommending}  propose a multi-task architecture, that makes a prediction for each objective and adopts as the loss a combined score using a combination function in the form of weighted multiplication. However, lacking the capability to constrain individual objectives, such a loss function may improve the combined score at the expense of severely jeopardizing some objectives, which is unacceptable for navigation service. 

{\bfseries Deep Learning for Combinatorial Optimization.} Deep learning has shown promising potential for solving combinatorial optimization problems. In \cite{vinyals2015pointer}, pointer network is proposed to learn approximate solutions to the travelling salesman problem. However, no feasible data-driven solution has been proposed so far to tackle the multi-objective routing problem in city-wide road networks.

\section{PRELIMINARIES\label{preliminaries}}
In this section, we provide key definitions, and outline the joint optimization problem of route planning.

{\bfseries Road Network}. A road network is a directed graph $G = (V, E)$, where $V$ is a vertex set and $E$ is an edge set. Each edge $e \in E$ is denoted by an ordered pair of vertices $(u, v) \in V \times V$. In particular, an edge could either refer to a road segment or a turn at intersection.


{\bfseries Route.} Given a routing request $q = (e_o$, $e_d)$,  a route $p$ between the pair is a sequence of edges $\{e_o = e_1, e_2, e_3, \ldots, e_{N - 1}, e_N = e_d\}$, where $e_i,  e_{i + 1}, \forall i \in [1, N - 1]$ are adjacent in the graph.

{\bfseries Route and Edge Attribute.} While most of route attributes are directly available from basic map data, such as distance, number of traffic lights, toll, etc., some attributes—for instance, deviation rate and travel time—are unknown in advance and thus need prediction. We build a separate estimator for each of these attributes. Let $c^{k},1 \le k \le K$ be all the route attributes considered in this work, and $c_p^k \in \mathbb R_{\ge 0}$ refer to the $k$th attribute value of route $p$. A similar convention holds for edges: $c_e^{k}, e \in E$ is just the edge weight.

{\bfseries Pareto Set.} Given an OD pair $q$, let $X_q$ be the set of all legitimate routes of $q$. Let $x, y \in X_q$ be two distinct routes. $x$ dominates $y$ if and only if $c_x^k \le c_y^k$ holds for all $k \in \{1, \ldots, K\}$ and the strict inequality holds at least once. Let $X_q^D = \{x \in X_q \, | \, \exists y \in X_q \, \text{such that} \, y \, \text{dominates} \, x\}$ be the set of dominated routes. Then $X_q^P = X_q - X_q^D$ is the set of non-dominated routes, or the Pareto-optimal routes. We call $X_q^P$ the Pareto set of $q$.

{\bfseries Multi-Objective Routing (MOR).} MOR recalls the Pareto set with respect to given objectives. Let $\boldsymbol{c^{\text{MOR}}} = (c^{o_1}, c^{o_2}, \ldots, c^{o_J})$ denote the $J$ objectives of MOR used in this paper.

{\bfseries Cost Function.} A cost function $f:E \to \mathbb R_{\ge 0}$ maps an edge $e \in E$ to a non-negative real value $c \in \mathbb R_{\ge 0}$. Given an OD pair $q$, the cost function uniquely defines the shortest path of $q$ in its own metrics. We use cost function and metrics interchangeably in the rest of the paper.

{\bfseries Route Preference.} We formalize route preference $\mathscr{P}$ as a task of constrained optimization
\begin{align}
\mathscr{P} &= \{h^0, \: (h^1, h^2, \ldots, h^L), \: (b^1, b^2, \ldots, b^L)\}
\end{align}
where $h^l \in \{c^1, c^2, \ldots, c^K\}, \, b^l \in \mathbb R$, $h^0$ denotes the objective attribute, $h^l, b^l, 1 \le l \le L$ denotes the constrained attributes and associated thresholds. To facilitate the following description, we use an index converter $\mathscr{C}$ to describe the correspondence between $h$ and $c$—that is, $h^{l} = c^{\mathscr{C}(l)}$.

{\bfseries Joint Optimization Problem of Route Planning.} Like classic recommender systems, route planning is an item-retrieval process with both a recall and a ranking stage. Specifically, provided a set of cost functions $F = \{f_1, f_2, \ldots, f_M\}$ and a routing request $q$, the recall module $R$ runs pathfinding algorithms to retrieve the paths $R(q)=\{p_1, p_2, \ldots, p_{M^\prime}\} \subset X_q $. A scoring function $g: X_q \to \mathbb R$ scores each path, and the highest-rated one $p_{\text{opt}}(q)= \argmax_{p \in R(q)}g(p)$ is recommended to the user. Let $Q$ denote a set of requests. Given a route preference $\mathscr{P}$, we formalize the joint optimization problem of route planning as
\begin{align}
\min_{F, g}\,&\frac{1}{|Q|}\sum\limits_{q \in Q}h^0_{{p_{\text{opt}}(q)}}\\
s.t. \quad &\frac{1}{|Q|}\sum\limits_{q \in Q}h^l_{{p_{\text{opt}}(q)}} \le b^l, \quad l = 1, \ldots, L \label{const}
\end{align}
where $F, g$ are learnable functions.

\section{METHODOLOGY\label{methodology}}
\subsection{Overall Architecture}
\begin{figure*}[t!] 
	\centering 
	\includegraphics[width=0.85\textwidth]{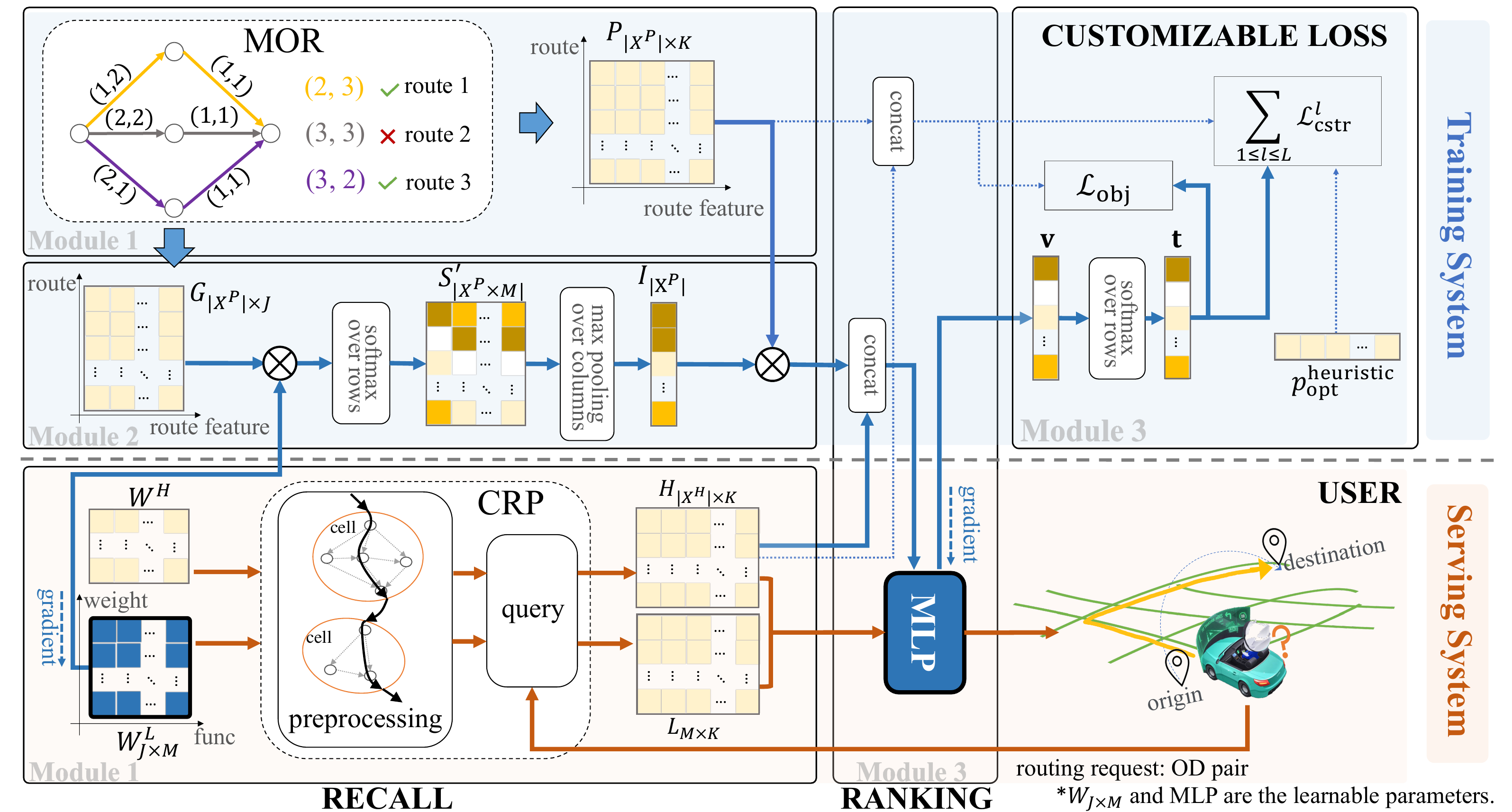} 
	\caption{The flowchart of the overall architecture.} 
	\label{fig:architecture} 
\end{figure*}

In this section, we describe the overall architecture of the learning framework, as illustrated in Figure \ref{fig:architecture}.

We first run pathfinding algorithms to generate the candidate routes for each request (module 1). Specifically, we apply CRP with the heuristic cost functions to retrieve the route set $X^H$ (heuristic route set), and multi-objective Dijkstra (MOD) to retrieve the Pareto set $X^P$. Using the attribute estimators inferred from historical trajectories, we transform $X^H, X^P$ into the feature matrices $ H_{|X^H| \times K}, P_{|X^P| \times K}$.

To learn the cost functions in $F$, we design module 2 to emulate shortest-path finding in a differentiable manner. First, we set all $f$ to be linear combinations of the $J$ objectives used in MOD—that is, $f_i(e)=\boldsymbol{c}_e^{\text{MOR}} \cdot \boldsymbol{w}_i^T, e \in E$. We then extract a submatrix $G_{|X^P| \times J}$ from $P$ with columns of $\boldsymbol{c}^{\text{MOR}}$. Next, we multiply $G$ by $W^L_{J \times M} = [ \boldsymbol{w}_1, \boldsymbol{w}_2, ..., \boldsymbol{w}_M ]$, the weight matrix of $F$, to obtain route costs under every metrics. At last, by applying softmax over the route dimension (rows) and max pooling over the metrics dimension (columns), we obtain a soft indicator vector $I_{|X^P|} $ of the routes recallable with $F$.

In module 3, we use MLP to score each route, and apply again the indicator technique to approximately pick up the feature vector of the top-rated route. This vector is fed into a loss function customized in line with the preference $\mathscr{P}$, generating the supervising signal.

We elaborate each of the modules in subsequent sections.

\subsection{Amap's Route-planning System}
To compute the shortest path, Amap uses the customizable route-planning algorithm (CRP) \cite{delling2017customizable}. To be specific, CRP has a two-stage preprocessing phase and a query phase. The road network is first partitioned into a few nested levels of cells with a minimum number of boundary (cross-cell) edges (metrics-independent stage). The cost of shortest path between every pair of boundary edges (called a “shortcut”) is next computed within each cell (metrics-dependent stage). Exploiting high-level shortcuts, the query algorithm takes much less steps to find the optimal route, resulting in two to three orders of magnitude of speed-up compared to plain Dijkstra. To further acquire alternative routes \cite{abraham2010alternative}, we use a bi-directional search in the query phase, which does not terminate until sufficient vertices have been doubly scanned (called “via vertex”). For every via vertex, the concatenation of the shortest paths origin-to-via-vertex and via-vertex-to-destination uniquely defines an alternative route.

Amap has several heuristic cost functions in-service. Depending on the route option, one or many of them are utilized to recall routes. In our experiment, we use two heuristic cost functions: “fastest in the free-flow period” and “avoid congestion”, as detailed in Table \ref{tab:cost_function}.

\subsection{Multi-Objective Routing and the Pareto Set}
As far as the objectives are concerned, the Pareto set is considered to include all valuable routes, therefore representing an upper bound achievable for a route recall system. We use as the objectives the five attributes that are most commonly concerned by digital map users: the distance, the live travel time, the toll, the number of traffic lights, and the number of driving maneuvers (turns). To compute the Pareto set $X^P$, we exploit the multi-objective Dijkstra algorithm \cite{delling2015round}, which has an exponential time complexity. 

\subsection{Feature Matrices of Candidate Routes}
In this work, two route attributes need prediction: the travel time and the deviation rate. To predict each attribute, we train a separate estimator with historical user trajectories. Specifics of the estimators are as follows:
\begin{itemize}
	\item Estimated time of arrival (ETA): Using real travel time as a label, we train a regression model with MSE loss to predict the travel time of any given route.
	\item Deviation rate estimator: Deviation rate is a categorical variable with two values: 1 if the user deviates from the top-ranked route, and 0 otherwise. We train a binary classifier to predict the deviation probability of any given route.
\end{itemize}
Using the above estimators, we obtain the feature matrices $ H \in \mathbb{R}_{\ge 0}^{|X^H| \times K}, P \in \mathbb{R}_{\ge 0}^{|X^P| \times K}$ for the heuristic set and the Pareto set. 

To prevent numerical issues and to ensure convergence, we perform the normalization on $H, P$ as follows: Given a matrix $X \in \mathbb{R}^{M \times N}$, let $x_{j}^{\text{max}} = \max_{1 \le i \le M}x_{ij}$. We transform $x_{ij}$ into $\frac{(x_{j}^{\text{max}} - x_{ij} + 1)}{x_{v}^{\text{max}}}$, where $v$ is a fixed index that corresponds to a feature with the largest numerical magnitude—distance in this paper.

\subsection{Differentiable Shortest Path Search}
We set all $f$ to be the weighted sum of the objectives used in MOD. Such a formulation not only eases the interpretation but also makes it possible to emulate shortest-path finding in a differentiable manner, as stated by the following theorem:

{\bfseries Theorem 1.}  Given $J$ objectives $\boldsymbol{c} = (c^{o_1}, c^{o_2}, \dots, c^{o_J})$ and an OD pair $q$, let $X_q^P$ denote the Pareto set with respect to $\boldsymbol{c}$. For an arbitrary weight vector $\boldsymbol{w} \in \mathbb{R}_{>0}^{1 \times J}$, the linear cost function $f(e)=\boldsymbol{c}_e \cdot \boldsymbol{w}$ recalls the shortest path $p_0 \in X_q^P, p_0 = \argmin_{p \in X_q^P}\boldsymbol{c}_p \cdot \boldsymbol{w}$.

{\bfseries Proof.} Let $X_q^D$ denote the set of dominated routes, and $X_q = X_q^P \cup X_q^D$. Given an arbitrary route $p^{\prime} \in X_q$, if $p^{\prime} \in X_q^P$, then $\boldsymbol{c}_{p_0} \cdot \boldsymbol{w}  = \min_{p \in X_q^P}\boldsymbol{c}_p \cdot \boldsymbol{w} \le \boldsymbol{c}_{p^{\prime}} \cdot \boldsymbol{w}$, concluding the proof. Otherwise, $p^{\prime} \in X_q^D$. So there must exist a route $p^{\prime\prime} \in X_q^P$ which dominates $p^{\prime}$—that is, $c_{p^{\prime\prime}}^k \le c_{p^{\prime}}^k$ holds for all $k \in \{1, \ldots, K\}$ and the strict inequality holds at least once. Clearly, $\boldsymbol{c}_{p_{\prime\prime}} \cdot \boldsymbol{w} - \boldsymbol{c}_{p_{\prime}} \cdot \boldsymbol{w} = \sum_{1 \le k \le K}(c_{p^{\prime\prime}}^k - c_{p^{\prime}}^k)w_k < 0$. Therefore, $\boldsymbol{c}_{p_0} \cdot \boldsymbol{w}  = \min_{p \in X_q^P}\boldsymbol{c}_p \cdot \boldsymbol{w} \le \boldsymbol{c}_{p^{\prime\prime}} \cdot \boldsymbol{w} < \boldsymbol{c}_{p^{\prime}} \cdot \boldsymbol{w}$. The proof is concluded.

According to Theorem 1, finding the shortest path of a cost function is equivalent to sorting the Pareto set by the same metrics. Following this idea, we first extract a submatrix $G \in \mathbb{R}^{|X^P| \times J}$ from $P$ with columns of $\boldsymbol{c}^{\text{MOR}}$—that is, $G_{ij} = P_{io_j}$. Let $W^L \in \mathbb{R}^{J \times M}$ denote the matrix representation of $F$, with the $j$th column representing the weight vector in $f_j \in F$, and $S \in \mathbb{R}^{|X^P| \times M} = G \times W^L$. $S_{ij}$ is the cost of route $i$ with respect to $f_j$. To identify the shortest route for each cost function, we apply softmax on $S$ over the rows and obtain the soft indicator matrix $S^{\prime} \in \mathbb{R}^{|X^P| \times M}$
\begin{align}
S_{ij}^{\prime} &= \dfrac{e^{S_{ij}/\tau_1}}{\sum\limits_{1 \le k \le |X^P|}e^{S_{kj}/\tau_1}} , \quad 1 \le i \le |X^P|, 1 \le j \le M
\end{align}
where the value of the shortest route in each metrics can vary from $\frac{1}{|X^P|}$ to unity, depending on the tunable scaling factor $\tau_1$. Since different cost functions may recall the same shortest route, we apply max pooling on $S^{\prime}$ over the columns (metrics dimension) to remove duplicates, obtaining a soft indicator vector $\boldsymbol{s}^{\prime\prime}  \in \mathbb R^{|X^P|}$:
\begin{align}
s_{i}^{\prime\prime} = \max_{1 \le j \le M}{S_{ij}^{\prime}}, \quad 1 \le i \le |X^P|
\end{align}

\subsection{Ranking Module}
Input of the ranking module is a concatenation of two route-feature matrices, $U = \left[\begin{smallmatrix} H\\P^{\prime} \end{smallmatrix}\right]$, where $H$ corresponds to the heuristic routes, and $P^{\prime}$ corresponds to the routes recalled with $F$. To obtain $P^{\prime}$, we approximately zero the rows in $P$ associated with the non-recallable routes of $F$, using the previous soft indicator vector:
\begin{align}
P_{ij}^{\prime} = P_{ij} \times s_{i}^{\prime\prime}, \quad 1 \le i \le |X^P|, 1 \le j \le M
\end{align}

Next, we use a multi-layer perceptron (MLP) to score each route and generate the rating vector $\boldsymbol{v}$ with $v_{i} = \text{MLP}(\boldsymbol{u}_i)$, where $\boldsymbol{u}_i \in \mathbb{R}^K$ is the $i$th row of $U$. By applying softmax on $\boldsymbol{v}$, we further acquire a soft indicator vector $\boldsymbol{t} \in \mathbb R^{1 \times (|X^H| + |X^P|)}$ of the top-ranked route:
\begin{align}
t_{i} = \dfrac{e^{v_i/\tau_2}}{\sum\limits_{1 \le i \le |X^H| + |X^P|}{e^{v_i/\tau_2}}}
\end{align}
At last, we obtain the attribute vector of the top-ranked route in two forms:
\begin{itemize}
	\item Approximate form $\boldsymbol{a} \in \mathbb R^{1 \times K}$: $\boldsymbol{a} = \boldsymbol{t} \times U$.
	\item Exact form $\boldsymbol{a}^\prime \in \mathbb R^{1 \times K}$: $\boldsymbol{a}^\prime$ is the $ i_{\text{opt}}$th row of $\left[\begin{smallmatrix} H\\P \end{smallmatrix}\right]$, where $ i_{\text{opt}} = \argmax_{1 \le i \le |X^H| + |X^P|}t_i$.
\end{itemize}

\subsection{Model Training}
{\bfseries Local Constraints.}
Section \ref{preliminaries} gives the formulation of route preference: an objective attribute plus a set of constrained attributes with thresholds. To avoid badcases, we further enforce the constraints to be locally valid across a three-dimensional parametric space of routing request $\mathbb{Z} = \mathbb{D} \times \mathbb{T} \times \mathbb{W}$, where $d \in \mathbb{D} = \mathbb{R}_{>0}$ denotes the great-circle distance of the OD pair, $t \in \mathbb{T} = \{t \in \mathbb{N} \, | \, 0 \le t < 24 \}$ denotes the hour index of the request time, and $\mathbb{W} = \{weekday, weekend\}$ denotes the weekday/weekend attribute of the request time. To be precise, the constraints hold in any continuous subspace $\mathbb{S} \subset \mathbb{Z}$.

{\bfseries Customizable Loss Function.}
To perform the constrained optimization, we propose the customizable loss function:
\begin{align}
\mathcal{L} &= \mathcal{L}_\text{obj} + \sum\limits_{1 \le l \le L}\mathcal{L}_\text{cstr}^l \label{eqn:loss}
\end{align}
The loss $\mathcal{L}_\text{obj}$ denotes the mean value of the objective attribute of the top-ranked route in the joint system (referred to as $p_\text{opt}^\text{joint}$):
\begin{align}
	\mathcal{L}_\text{obj} &= \dfrac{1}{|\mathbb{S}|}\sum\limits_{\mathbb{S}} a_{\mathscr{C}(0)}
\end{align}
The other loss $\mathcal{L}_\text{cstr}^l$ yields a penalty when the corresponding constraint gets violated:
\begin{align}
\mathcal{L}_\text{cstr}^l &= \lambda_l \times \max(0, \mathcal{H}^l - b^l) \\
\end{align}
with
\begin{align}
\mathcal{H}^l &= \dfrac{1}{|\mathbb{S}|}\sum\limits_{\mathbb{S}} t_{i_{\text{opt}}} \times a_{\mathscr{C}(l)}^{\prime}\\
b^l &= \dfrac{1}{|\mathbb{S}|}\sum\limits_{\mathbb{S}} t_{i_{\text{opt}}} \times {h_{p_{\text{opt}}^{\text{heuristic}}}^l} \times (1 + o_l)
\end{align}
where $p_{\text{opt}}^{\text{heuristic}}$ refers to the top-ranked route in Amap's route-planning system, $\mathcal{H}^l$ denotes the mean value of the constrained attribute of $p_\text{opt}^\text{joint}$, $b^l$—the threshold—denotes the mean value of the constrained attribute of $p_{\text{opt}}^{\text{heuristic}}$ scaled by a factor of  $1 + o_l$, and $\lambda_l > 0$ are hyperparameters controlling the magnitude of the constraints. In essence, $\mathcal{L}_\text{cstr}^l$ constrain the attribute value of the joint system to be less than $1 + o_l$ times that of the existing system.

{\bfseries Batch Sampling Method.}
To enforce local constraints, we propose a novel two-stage batch sampling method as follows. Let $p(z_0)$ denote the request density at $z_0 = (d_0, t_0, w_0) \in \mathbb{Z}$. In each iteration, we first randomly sample a batch $\mathbb{B}_\text{obj}$. We then pick up an anchor point $z_c = (d_c, t_c, w_c) \in \mathbb{Z}$ based on the probability $p$. Next, we randomly sample a sub-batch of requests $\mathbb{B}_\text{cstr} \subset \mathbb{B}_\text{obj}$, within a window centered on $z_c$, $\{(d, t, w) \, | \, d_c - \Delta{d} \le d \le d_c + \Delta{d}, \, t_c - \Delta{t} \le t \le t_c + \Delta{t}, \, w = w_c \}$, where $\Delta d = 5 \, \text{km}$ and $\Delta c = 2 \, \text{hours}$ are constants. We calculate $\mathcal{L}_\text{obj}$ on $\mathbb{B}_\text{obj}$ and $\sum\limits_{1 \le l \le L}\mathcal{L}_\text{cstr}^l$ on $\mathbb{B}_\text{cstr}$.

{\bfseries Hyperparameter Optimization.}
Proper choice of the hyperparameters $\lambda_l$ are critical for achieving a converged solution that satisfies all inequality constraints: a large $\lambda_l$ may lead to oscillation during the iteration, yet a small $\lambda_l$ may cause violation of the constraints. To tackle this issue, we use scenariolized $\lambda_{l, s}$, where $s$ denotes a point on a grid in $\mathbb{Z}$, initialize all $\lambda_{l, s}$ as random numbers between 0 and 1, and update them after each iteration as follows:
\begin{equation}
\lambda_{l, s}^{n+1} = \begin{cases}
\lambda_{l, s}^{n} + \Delta_\lambda &, \mathcal{H}^l > b^l
\\
\text{max}(0, \lambda_{l, s}^{n} - \Delta_\lambda)&, \text{otherwise}
\end{cases}
\label{eq:wd}
\end{equation}
where $n$ denotes the index of iteration and $\Delta_\lambda = 0.01$ is a constant.

\section{EXPERIMENT\label{experiment}}
In this section, we first describe the datasets, methods, implementation details, and evaluation metrics. Then we show the effectiveness of the customizable loss function and the joint learning framework.

\subsection{Datasets}

We conduct experiments based on two datasets. One dataset contains routing requests and GPS trajectories of anonymous users of Amap in Beijing and Shanghai from November 1, 2020 to November 7, 2020. Each request has five fields $q = (t, x_o, y_o, x_d, y_d)$, where $t$ is the timestamp when the request was received by Amap; $x_o, y_o$ are the longitude, latitude of the origin; and  $x_d, y_d$ are the longitude, latitude of the destination. The other dataset contains information on the road networks, including the adjacency relation and attributes of the edges. Table \ref{tab:dataset} shows the statistics of the trajectories and road networks. We randomly split the data into 80\% for training and 20\% for testing.

\begin{table}[htbp]
	\centering
	\caption{Statistics of the datasets.}
	\begin{tabular}{ccrr}
		\toprule
		Dataset &   Attribute     & Beijing & Shanghai \\
		\midrule
		\multirow{8}[2]{*}{\makecell[t]{User \\ trajectory}} 
		& Num. of records & 1.38M & 1.22M \\
		& travel time & 1584s & 1604s \\
		& travel distance & 14.4km & 14.2km \\
		& length of congestion & 472m & 262m \\
		& Num. of traffic lights & 8.3 & 12.7 \\
		& Num. of turns  & 9.2 & 10.2 \\
		& toll & 2.15\textyen & 1.70\textyen \\
		\midrule
		\multirow{3}[2]{*}{\makecell[t]{Road \\ network}} 
		& Num. of roads & 2.09M & 2.21M \\
		& Num. of toll roads & 5.68k & 4.02k \\
		& Num. of traffic lights & 9.71k & 14.8k \\
		& Num. of turns & 4.21M & 4.53M \\
		\bottomrule
	\end{tabular}%
	\label{tab:dataset}%
\end{table}%

\subsection{Methods}

Methods compared in the experiment are detailed as follows:

{\bfseries Benchmark Methods.}
\begin{itemize}
	\item H-SD: Routes in H-SD are recalled with the heuristic cost functions, and ranked according to distance.
	\item H-ST: Routes in H-ST are recalled with the heuristic cost functions, and ranked according to ETA.
	\item H-Lambda: In the industry of digital map and navigation, deviation rate is widely adopted as the measure of route quality. To lower the deviation rate, a ranking model with listwise loss function is commonly used, as elaborated below. Let $S_q$ denote the set of routes recalled with respect to request $q$, and $t_q$ denote user trajectory. A route $p \in S_q$ is labelled as 1 if it completely coincides with $t_q$; 0 otherwise. Sorted according to such a label, $S_q$ becomes an ordered list, which serves as the ground truth of a ranking model. In H-Lambda, we train a LambdaMART model \cite{Kosiur01} to rank the heuristic routes. We set the number of trees at 1150, feature sample rate at 0.5, and data sample rate at 0.5.
\end{itemize}

{\bfseries Customizable and Jointly Optimized Route Planning.}
\begin{itemize}
	\item H-$\text{CJRP}_{\text{R}}$: H-$\text{CJRP}_{\text{R}}$ denotes a route-planning system comprising the heuristic routes, and a ranking model (MLP) trained with the customized loss function. We use two layers for the MLP, and ELU as the activation. The hidden unit of each layer is 64.
	\item H-$\text{CJRP}_{\text{R}}^\prime$: H-$\text{CJRP}_{\text{R}}^\prime$ is the same as H-$\text{CJRP}_{\text{R}}$, except for using the plain single-stage random batch sampling with fixed $\lambda$.
	\item E-$\text{CJRP}_{\text{R}}$: In CRP, properties of the alternative routes depend on three control parameters $(\alpha, \beta, \gamma)$, where $\alpha$ denotes the maximum allowed stretch, $\beta$ denotes the maximum allowed route sharing, and $\gamma$ denotes the threshold guaranteeing local optimality. We relax the control parameters to recall additional alternative routes and denote such an enlarged route set by $S_E$. E-$\text{CJRP}_{\text{R}}$ denotes a route-planning system comprising the enlarged set of routes, and a ranking model (MLP) trained with the customized loss function, where MLP has the same structure as in H-$\text{CJRP}_{\text{R}}$.
	\item M-$\text{CJRP}_{\text{R}}$: M-$\text{CJRP}_{\text{R}}$ denotes a route-planning system comprising the union of the heuristic route set and the Pareto set, and a ranking model (MLP) trained with the customized loss function, where MLP has the same structure as in H-$\text{CJRP}_{\text{R}}$.
	\item $\text{CJRP}_{\text{J}}$: $\text{CJRP}_{\text{J}}$ denotes a route-planning system comprising the union of the heuristic routes and the routes recalled with the learned cost functions, and a ranking model trained with the customized loss function, where MLP has the same structure as in H-$\text{CJRP}_{\text{R}}$.
\end{itemize}

Among the CJRP family, $\text{CJRP}_{\text{J}}$, which learns a set of cost functions to compensate the inefficiency of the existing heuristic cost functions, achieves the best balance between route quality and query response time.

\subsection{Implementation Details}
In Table \ref{tab:features}, we show the representative features of the ranking model, which are shared across all methods. In CRP, we set the control parameters as $\alpha = 0.85$, $\epsilon = 0.3$, and $\gamma = 0.2$ for the heuristic route set, and $\alpha = 0.9$, $\epsilon = 0.3$, and $\gamma = 0.1$ for the enlarged route set. In CJRP models, all $\lambda$s controlling the magnitude of the constraints are initialized as random numbers between 0 and 1. To train the CJRP models, we use Adam optimizer with initial learning rate 0.001 and decay rate 0.98. The ETA model is implemented with scikit-learn; the deviation rate model and H-Lambda are implemented with LightGBM; the neural network models are implemented with TensorFlow. The training and inference of all models are conducted on seven servers each with 512GB RAM and 96 cores.

\begin{table}[htbp]
	\centering
	\caption{Representative features in the ranking model.}
	\begin{tabular}{p{0.30\columnwidth}|p{0.58\columnwidth}}
		\toprule
		Feature Type &   Features  \\
		\midrule
		Temporal & hour, day of week, holiday, etc. \\
		\midrule
		Spatial & city, POI category, etc. \\
		\midrule
		Static attribute & distance, length of freeways, free-flow travel time, number of traffic lights, number of driving maneuvers, toll, etc.\\
		\midrule
		Dynamic attribute & live travel time, ETA, length of traffic congestion at the request time, deviation rate, etc. \\
		\midrule
		Badcase attribute & local detour, length of the alleyways, frequent switching between main road and auxiliary road, etc. \\
		\bottomrule
	\end{tabular}%
	\label{tab:features}%
\end{table}%

\subsection{Evaluation Metrics\label{metrics}}
We use the following four route preferences\footnote{Without loss of generality, we simplify the constraints from those in-service.} to evaluate model performance:
\begin{itemize}
	\item Regular: It recommends a route with the lowest deviation rate and tolerable degradation of other attributes. The objective attribute is deviation rate. The constrained attributes (threshold offsets $o_l$) are ETA (0.5\%), distance (0.5\%), and length of congested roads (0\%).
	\item Fastest: It recommends a route with the shortest travel time and tolerable degradation of other attributes. The objective attribute is ETA. The constrained attributes (threshold offsets $o_l$) are distance (5\%), number of traffic lights (0\%), and length of alleyways (0\%).
	\item Avoid tolls: It recommends a route with less toll roads than the regular route. The objective attribute is toll. The constrained attributes (threshold offsets $o_l$) are ETA (5\%) and distance (5\%).
	\item Economic: This option is provided for ride-hailing users. It recommends a route with the lowest order price and tolerable degradation of other attributes. The real pricing formula of an order is fairly complicated; we use a simplified version instead in this paper: $\text{price (\textyen)} =  3.15 \times \text{distance (kilometer)} + 0.7 \times \text{ETA (minute)} + \text{toll (\textyen)}$. The objective attribute is order price. The constrained attributes (threshold offsets $o_l$) are ETA (10\%) and distance (10\%).\\
\end{itemize}
The above implementation of these preferences are determined based on study of user feedback. We use the corresponding objective attribute and average response time (ART) as the evaluation metrics for each preference.

\subsection{Performance Comparison}
{\bfseries Effectiveness in Customizing Route Attributes.}
We conduct experiments on the regular preference to validate the effectiveness of H-$\text{CJRP}_{\text{R}}$ in customizing route attributes.

\begin{figure}[h] 
	\centering 
	\includegraphics[width=0.9\linewidth]{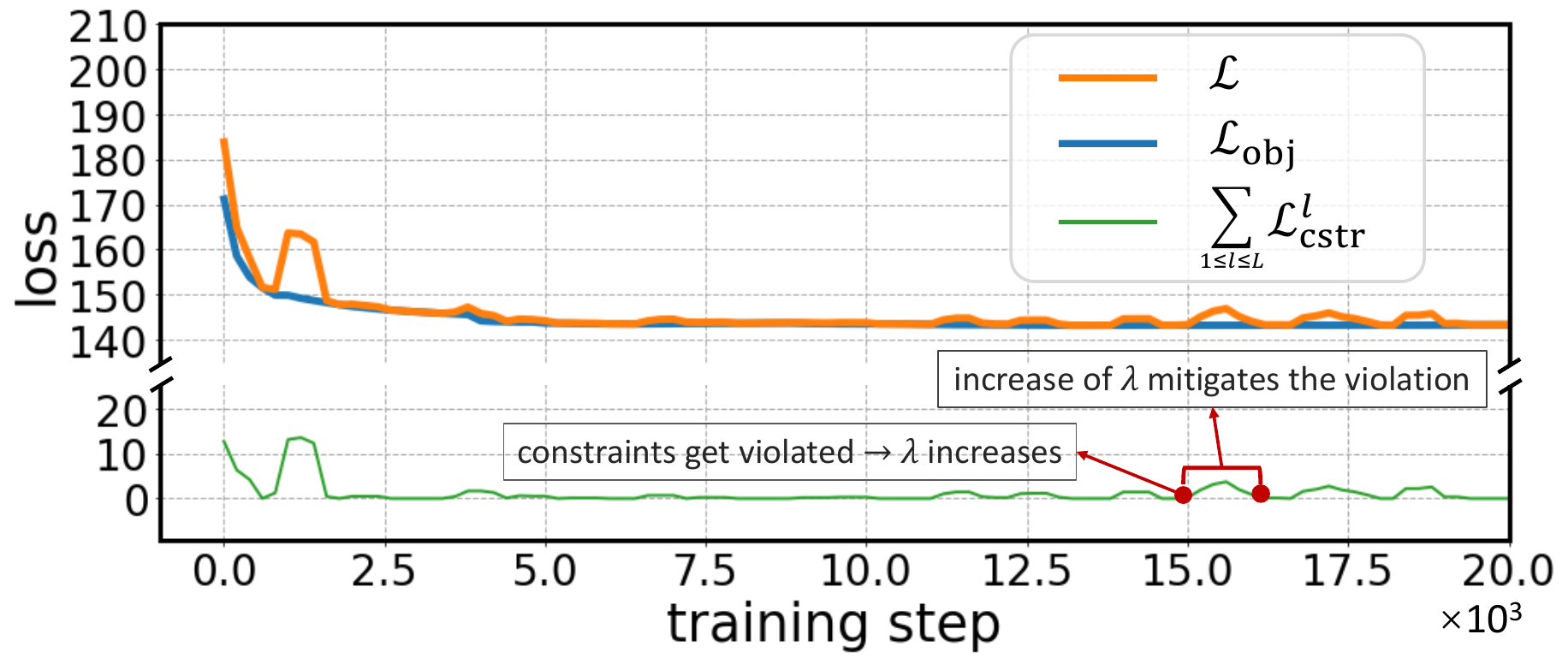} 
	\caption{Training loss curve of H-$\text{CJRP}_{\text{R}}$.} 
	\label{fig:loss_curve} 
\end{figure}

\begin{figure}[h] 
	\centering 
	\includegraphics[width=0.9\linewidth]{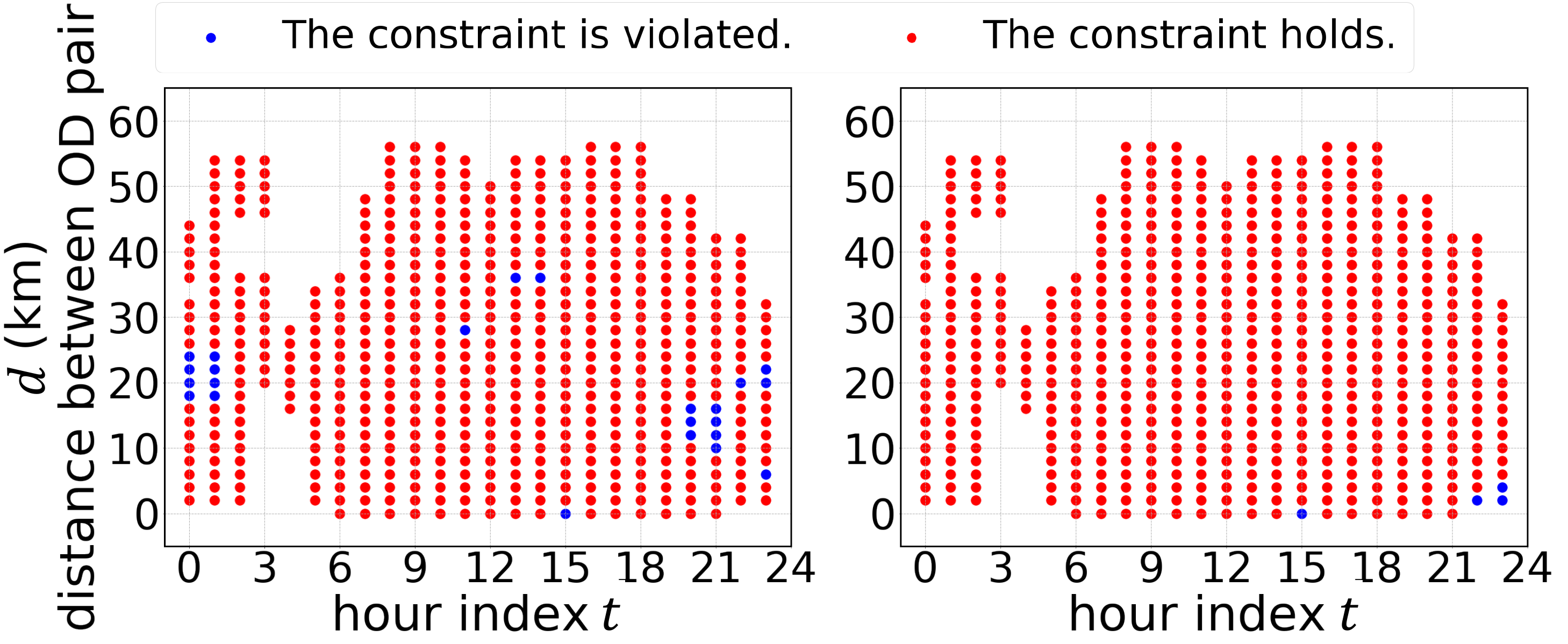} 
		\caption{Comparison of $\text{CJRP}^\prime$(left) and $\text{CJRP}$(right) on local constraint on the length of congested roads. Each point in the graph represents a distinct routing scenario. Clearly, much fewer scenarios violate the constraint in $\text{CJRP}$ than in $\text{CJRP}^\prime$.}
	\label{fig:constraint}
\end{figure}

We train H-Lambda and H-$\text{CJRP}_{\text{R}}$, with the regular loss function as detailed in Section \ref{metrics}. The training loss curve of H-$\text{CJRP}_{\text{R}}$ is shown in Figure \ref{fig:loss_curve}. Table \ref{tab:regular_offline} shows the mean values of the objective variables and constrained variables, where the deviation rate and ETA are predicted using estimators described in Section \ref{preliminaries}. In line with the customized loss function, H-$\text{CJRP}_{\text{R}}$ keeps ETA, distance, and length of congested roads within the imposed constraints, and achieves a significant improvement in deviation rate over H-Lambda. Also, all constraints are ensured locally valid across the parametric space of requests, as illustrated in Figure \ref{fig:constraint}. 

\begin{figure}[h]
	\centering
	\begin{minipage}[t]{1\linewidth}
		\begin{subfigure}[t]{0.48\textwidth}
			\centering
			\includegraphics[width=\linewidth]{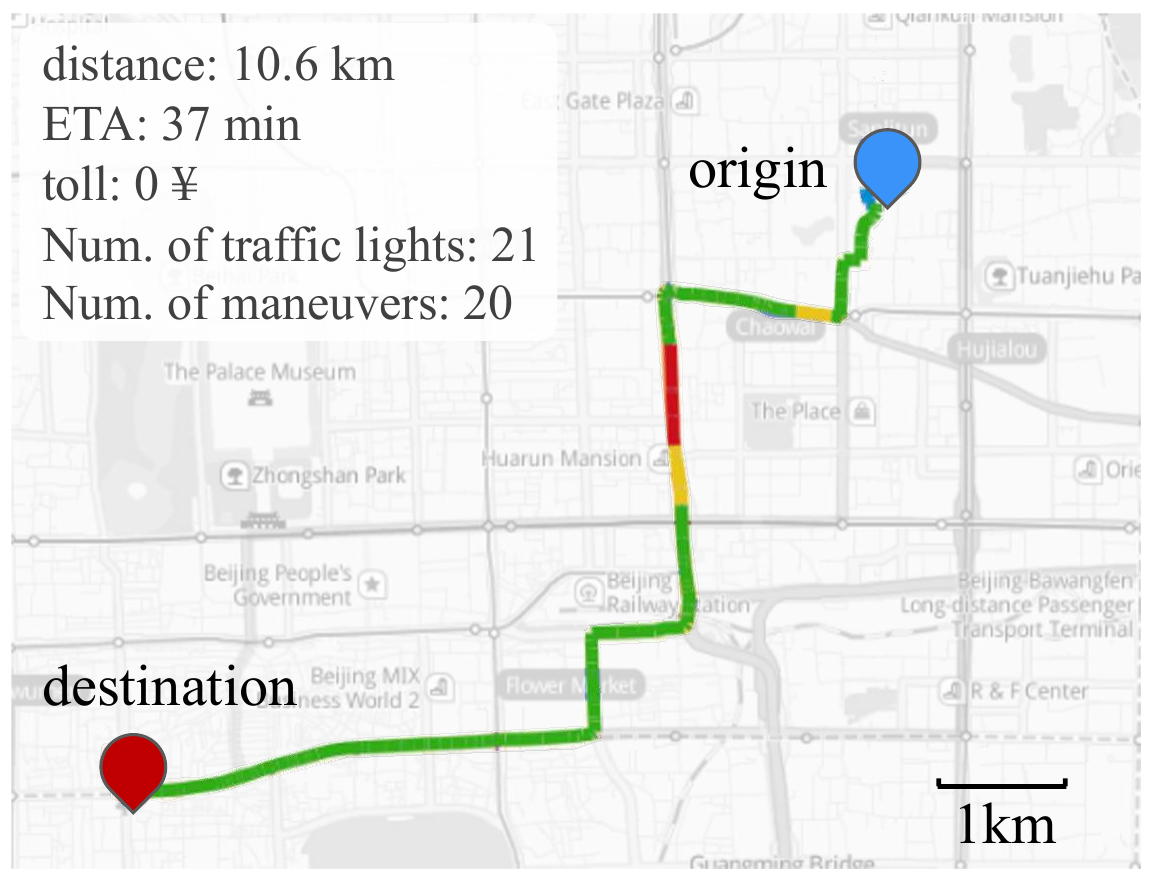} 
			\label{fig:libra_case_gray_left}
		\end{subfigure}
		\hfill
		\begin{subfigure}[t]{0.48\textwidth}
			\centering
			\includegraphics[width=\linewidth]{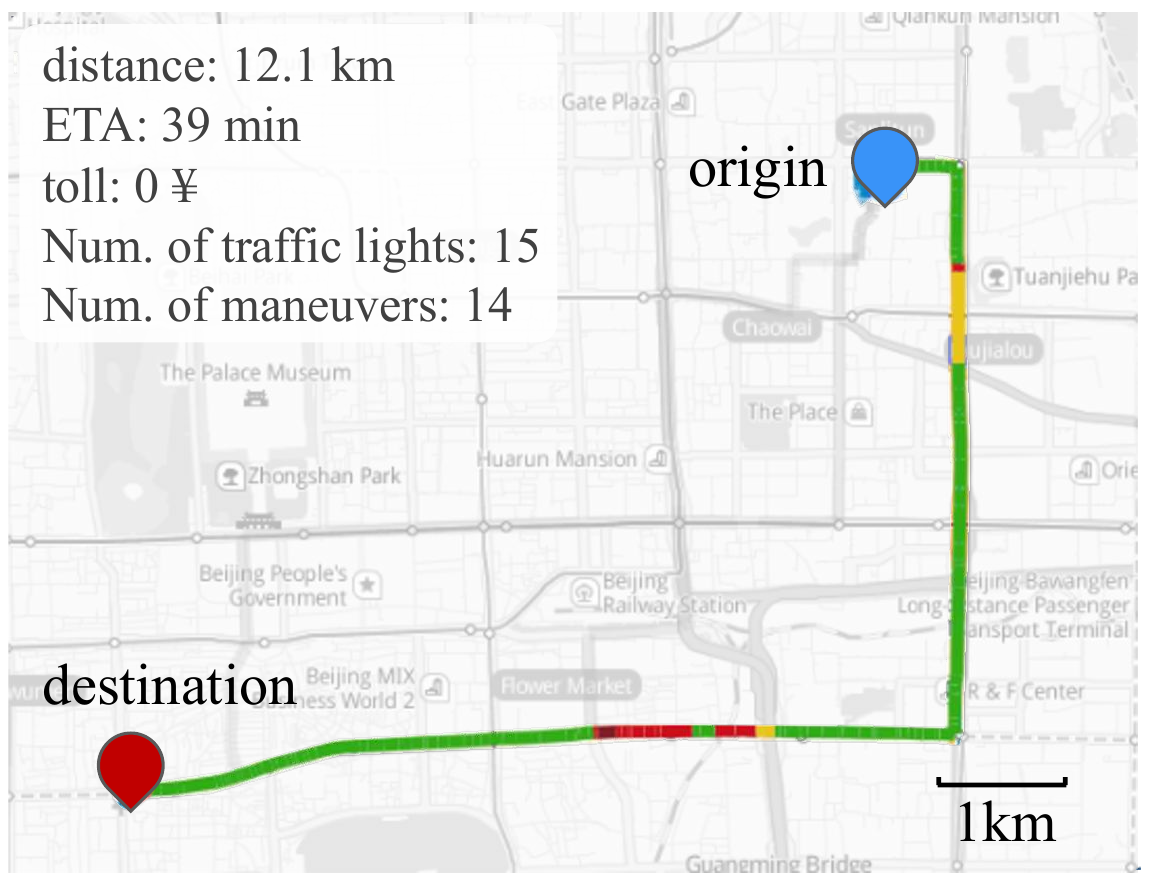} 
			\label{fig:libra_case_gray_right}
		\end{subfigure}
	\end{minipage} 
	\caption{Case study: H-Lambda (left) versus H-$\text{CJRP}_{\text{R}}$ (right) on the regular preference.}
	\label{fig:case-ranking}
\end{figure}

We further deploy the two models in Amap's online routing service in Beijing and Shanghai, from December 19 to December 21. As A/B testing, we randomly allocate the real-time routing requests to the two models. As shown in Table \ref{tab:regular_online}, H-$\text{CJRP}_{\text{R}}$ satisfies the constraint on travel time, and achieves a decent improvement in deviation rate over H-Lambda. Notice that the improvement in online evaluation is smaller compared to its offline counterpart. This is attributed to the inadequate predicative capability of the deviation rate estimator, which can be considerably strengthened by including stronger discriminative features, such as the individualized characteristics. To showcase the effect of H-$\text{CJRP}_{\text{R}}$, we provide an example in Figure \ref{fig:case-ranking}. Compared to the recommended route in H-Lambda, the top-ranked route in H-$\text{CJRP}_{\text{R}}$ has fewer driving maneuvers and traffic lights, looks smoother, and is much easier to follow. Intuitively, this route sacrifices a few minutes and kilometers in exchange for a lower deviation rate.

\begin{table}[htbp]
	\centering
	\caption{Comparisons between H-Lambda and H-$\text{CJRP}_{\text{R}}$ on the regular preference, based on offline evaluation.}
	\begin{tabular}{c|rr|rr}
		\toprule 
		Dataset & \multicolumn{2}{c|}{Beijing} & \multicolumn{2}{c}{Shanghai} \\
		\midrule
		Model & H-Lambda & H-$\text{CJRP}_{\text{R}}$ & H-Lambda & H-$\text{CJRP}_{\text{R}}$ \\
		\midrule
		deviation rate & 33.9\% & 32.2\% & 35.5\% & 33.9\% \\
		ETA & 1830 s & 1811 s & 1802 s & 1789 s \\
		distance & 18148 m & 18176 m & 17136 m & 17122 m \\
		congestion & 412 m & 393 m & 197 m & 190 m \\
		\bottomrule
	\end{tabular}%
	\label{tab:regular_offline}%
\end{table}%

\begin{table}[htbp]
	\centering
	\caption{Comparisons between H-Lambda and H-$\text{CJRP}_{\text{R}}$ on the regular preference in an online A/B testing.}
	\begin{threeparttable}
		\begin{tabular}{c|rr|rr}
			\toprule 
			Dataset & \multicolumn{2}{c|}{Beijing} & \multicolumn{2}{c}{Shanghai} \\
			\midrule
			Model & H-Lambda & H-$\text{CJRP}_{\text{R}}$ & H-Lambda & H-$\text{CJRP}_{\text{R}}$ \\
			\midrule
			deviation rate & 36.1\% & 35.3\% & 38.9\% & 38.2\% \\
			travel time & 1913 s & 1924 s & 2031 s & 2021 s \\
			\bottomrule
		\end{tabular}%
	\end{threeparttable}%
	\label{tab:regular_online}
\end{table}%

\begin{figure}[h]
	\centering
	\begin{minipage}[t]{1\linewidth}
		\begin{subfigure}[t]{0.48\textwidth}
			\centering
			\includegraphics[width=\linewidth]{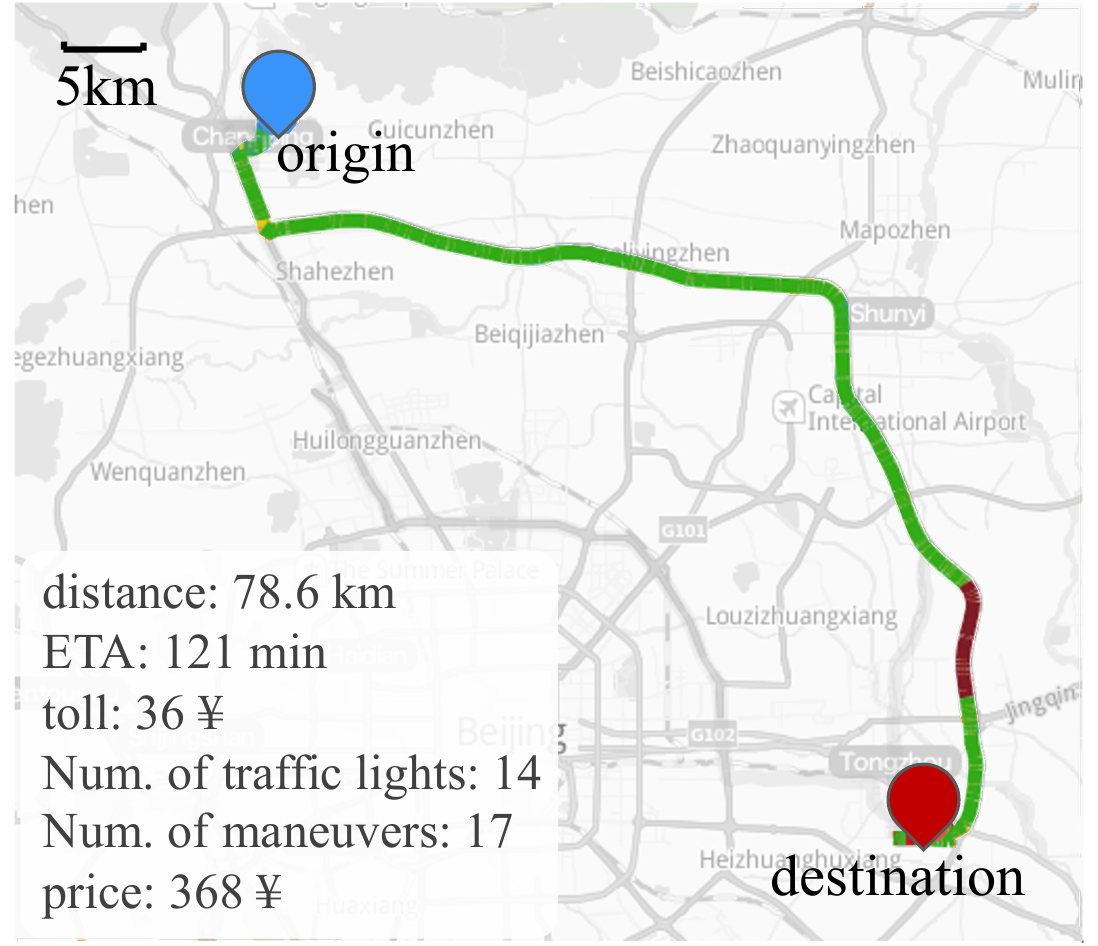} 
			\label{fig:libra_case_gray_left}
		\end{subfigure}
		\hfill
		\begin{subfigure}[t]{0.48\textwidth}
			\centering
			\includegraphics[width=\linewidth]{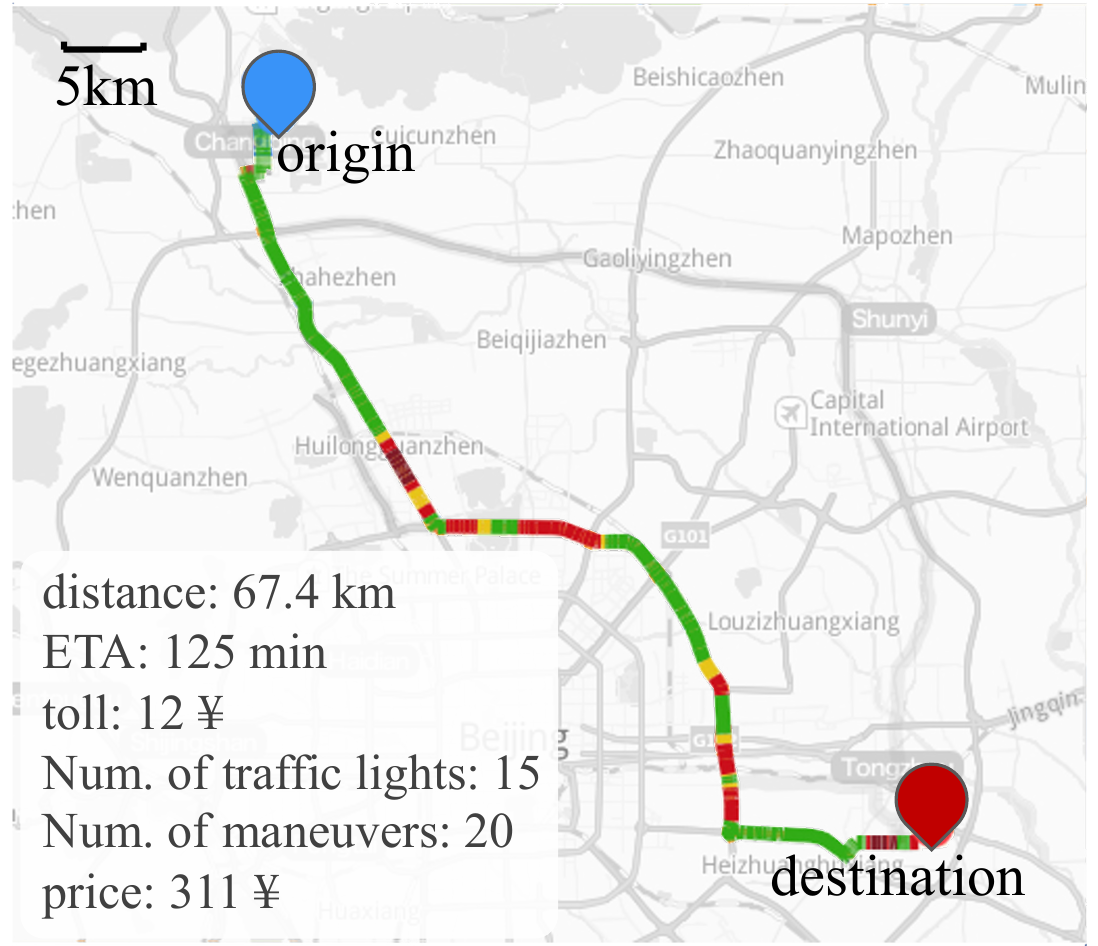} 
			\label{fig:libra_case_gray_right}
		\end{subfigure}
	\end{minipage} 
	\caption{Case study: H-$\text{CJRP}_{\text{R}}$ (left) versus $\text{CJRP}_{\text{J}}$ (right) on the economic preference. }
	\label{fig:case-cf}
\end{figure}

{\bfseries Customizability.}
We compare the offline performance of the CJRP family and the benchmarks on three other common preferences, as listed in Section \ref{metrics}. Table \ref{tab:offline} show the results. For every preference, all CJRP models significantly outperform the benchmarks, verifying the customizability of the proposed framework.

\begin{table*}[htbp]
	\centering
	\caption{Comparisons on various route preferences in Beijing and Shanghai, based on the offline evaluation. The attribute values of H-Lambda are the original numbers. To better contrast the performance, the attribute values of other models are displayed as the difference between the original numbers and those of H-Lambda. Numbers in the brackets are the corresponding percentage differences. We set $M = 5$ (number of learnable cost functions) in $\text{CJRP}_{\text{J}}$.}
	 \begin{threeparttable}[t]
	\begin{tabular}{c|llll|lll|lll}
		\toprule 
		&  \multicolumn{4}{c}{Fastest} &  \multicolumn{3}{|c}{Avoid tolls} &  \multicolumn{3}{|c}{Economic} \\
		\midrule
		\multirow{2}[2]{*}{Model} 
		& obj & cstr-1 & cstr-2 & cstr-3 & obj & cstr-1 & cstr-2 & obj & cstr-1 & cstr-2 \\
		\cmidrule{2-11}
		& ETA & distance &  traffic light  & alleyway & toll & ETA & distance & price & ETA & distance \\
		\midrule
		H-Lambda & 1280s & 10824m & 9.7  & 9.0m & 0.83\textyen  &1280s & 10824m & 49.86\textyen & 1280s & 10824m \\
		\midrule
		H-ST & -39s& 353m (3\%) & -0.3 (-3\%) & 1.4m (16\%) & 0.26\textyen & -39s (-3\%) & 353m (3\%) & 0.93\textyen & -39s (-3\%) & 353m (3\%) \\		
		H-SD & 118s & -606m (-6\%) & 2.9 (30\%) & 1.6m (18\%) & -0.23\textyen & 118s (9\%) & -606m (-6\%) & 0.76\textyen & 118s (9\%) & -606m (-6\%) \\		
		\midrule
		H-$\text{CJRP}_{\text{R}}$ &-37.9s& 361m (3\%) & -0.3 (-3\%) & -0.1m (-1\%) &  -0.48\textyen & 46s (4\%) & 207m (2\%) & -1.42\textyen & 32s (3\%) & -485m (-4\%) \\
		E-$\text{CJRP}_{\text{R}}$  &-41.5s& 414m (4\%) & -0.2 (-2\%) & -0.2m (-2\%) &  -0.50\textyen & 46s (4\%) & 288m (3\%) &-1.56\textyen & 35s (3\%) & -528m (-5\%) \\
		$\text{CJRP}_{\text{J}}$  &-64.5s& 359m (3\%)  & -1.0 (-10\%) & -0.3m (-3\%) &  -0.64\textyen & 56s (4\%) & 290m (3\%) & -1.87\textyen & 37s (3\%) & -615m (-6\%) \\
		M-$\text{CJRP}_{\text{R}}$ &-69.1s& 325m (3\%) & -1.1 (-11\%) & -0.1m (-1\%) &  -0.65\textyen & 59s (5\%) & 124m (1\%) & -2.02\textyen & 30s (2\%) & -634m (-6\%) \\
		\bottomrule
	\end{tabular}%
\end{threeparttable}%
	\label{tab:offline}%
\end{table*}%

\begin{table}[htbp]
	\centering
	\caption{The weights of the cost functions. H1 and H2 denote the two heuristic cost functions: “fastest in the free flow period” and “avoid congestion.” L1 to L5 denote the learned cost functions of $\text{CJRP}_{\text{J}}$ for the economic preference. Every cost function is a linear combination of the five attributes, and the values are the corresponding weights.}
	\begin{tabular}{p{0.36\linewidth}|cc|ccccc}
		\toprule
		Attribute & H1 & H2 & L1 & L2 & L3 & L4 & L5\\
		\midrule
		free-flow travel time (s) & 14 & 0 & 56 & 5 & 0 & 0 & 0 \\		
		distance (m) & 1 & 1 & 1 & 1 & 1 & 1 & 1\\
		live travel time (s) & 0 & 14 & 3 & 0 & 1 & 2 & 0\\ 
		traffic light & 114 & 114 & 517 & 0 & 0 & 0 & 0\\
		maneuver & 46 & 646 & 2504 & 397 & 0 & 0 & 0\\
		\bottomrule
	\end{tabular}%
	\label{tab:cost_function}%
\end{table}%

\begin{table}[htbp]
	\centering
	\caption{Comparison of the average response time}
	\begin{tabular}{c|c|c|c}
		\toprule 
		Model & Num. of routes & $\text{ART}_\text{recall}$ & $\text{ART}_\text{ranking}$ \\
		\midrule
		H-Lambda & 7 & 196 ms & 98 ms \\		
		\midrule
		H-$\text{CJRP}_{\text{R}}$ & 7 & 196 ms & 97 ms \\
		E-$\text{CJRP}_{\text{R}}$ & 28 & 228 ms & 319 ms  \\
		$\text{CJRP}_{\text{J}}$ & 12 & 222 ms & 160 ms \\
		M-$\text{CJRP}_{\text{R}}$ & 48 & 15793 ms & 603 ms \\
		\bottomrule
	\end{tabular}%
	\label{tab:recall}%
\end{table}%

\begin{figure}[t!] 
	\centering 
	\includegraphics[width=0.45\textwidth]{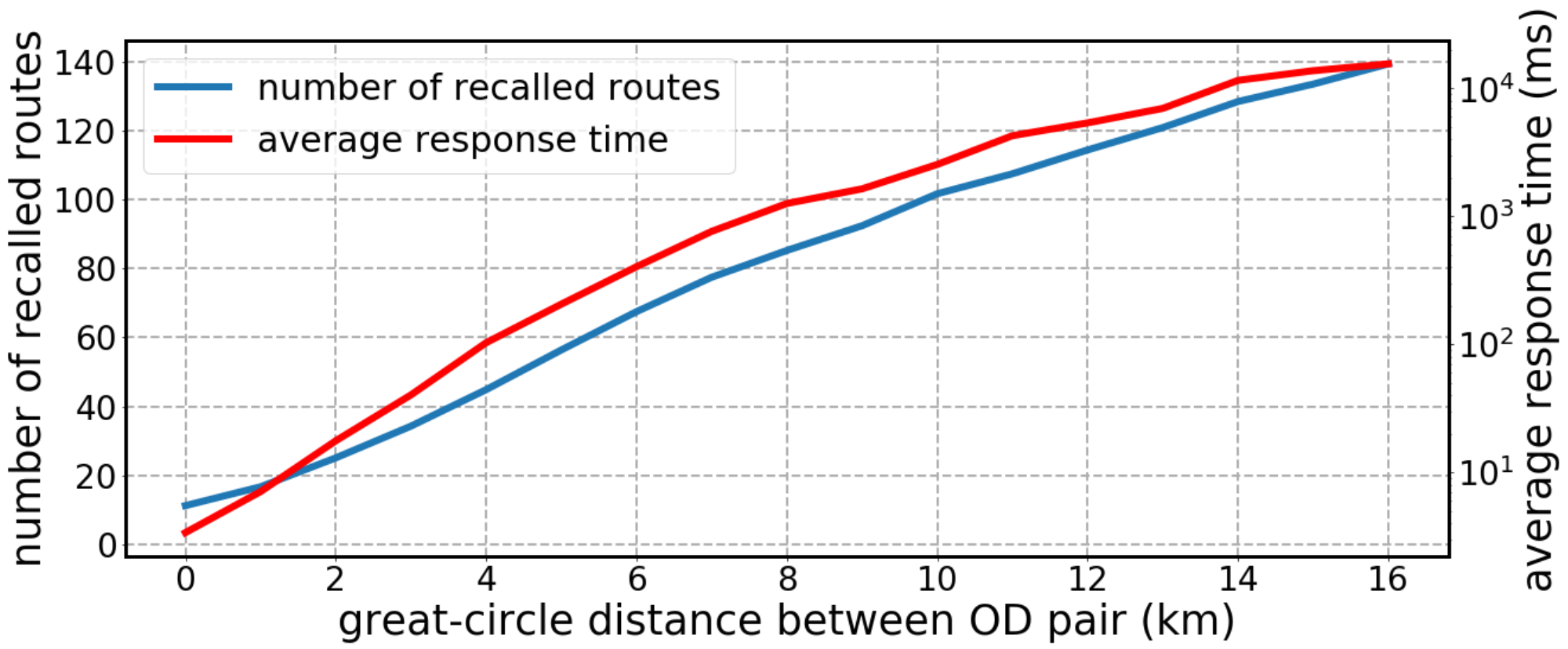} 
	\caption{Exponential time complexity of MOD.} 
	\label{tab:mod} 
\end{figure}

{\bfseries Effectiveness of the Learned Cost Functions.}
We further compare the performance among the CJRP models, which differ only in the recall stage. First, $\text{CJRP}_{\text{J}}$ considerably outperforms H-$\text{CJRP}_{\text{R}}$. 26.3\% of the top-ranked routes in $\text{CJRP}_{\text{J}}$ are solely recallable with the learned cost functions, suggesting the inefficiency of the existing heuristic cost functions. We show an example from the economic preference in Figure \ref{fig:case-cf}. The top-ranked route in $\text{CJRP}_{\text{J}}$, which is recalled with a learned cost function—L2 as shown in Table \ref{tab:cost_function}, has a minor increase in ETA, but a 15\% lower price. Such an improvement is attributed to the optimality of the learned cost function towards the preference: the weight ratio of travel time to distance in L2 is 5, about the same as that—3.7— in the pricing formula. Secondly, $\text{CJRP}_{\text{J}}$, though with fewer routes recalled, substantially outperforms E-$\text{CJRP}_{\text{R}}$, validating the effectiveness of the optimality of the learned cost functions. Furthermore, as mentioned above, multi-objective Dijkstra has exponential time complexity, which limits its application in real-time services. Figure \ref{tab:mod} illustrates this point on our dataset. Compared to M-$\text{CJRP}_{\text{R}}$, $\text{CJRP}_{\text{J}}$ achieves a similar level of performance with a much smaller and practical response time as shown in Table \ref{tab:recall}, demonstrating the superiority of the proposed framework in large-scale real-world application.


\begin{figure}[H] 
	\centering 
	\includegraphics[width=0.45\textwidth]{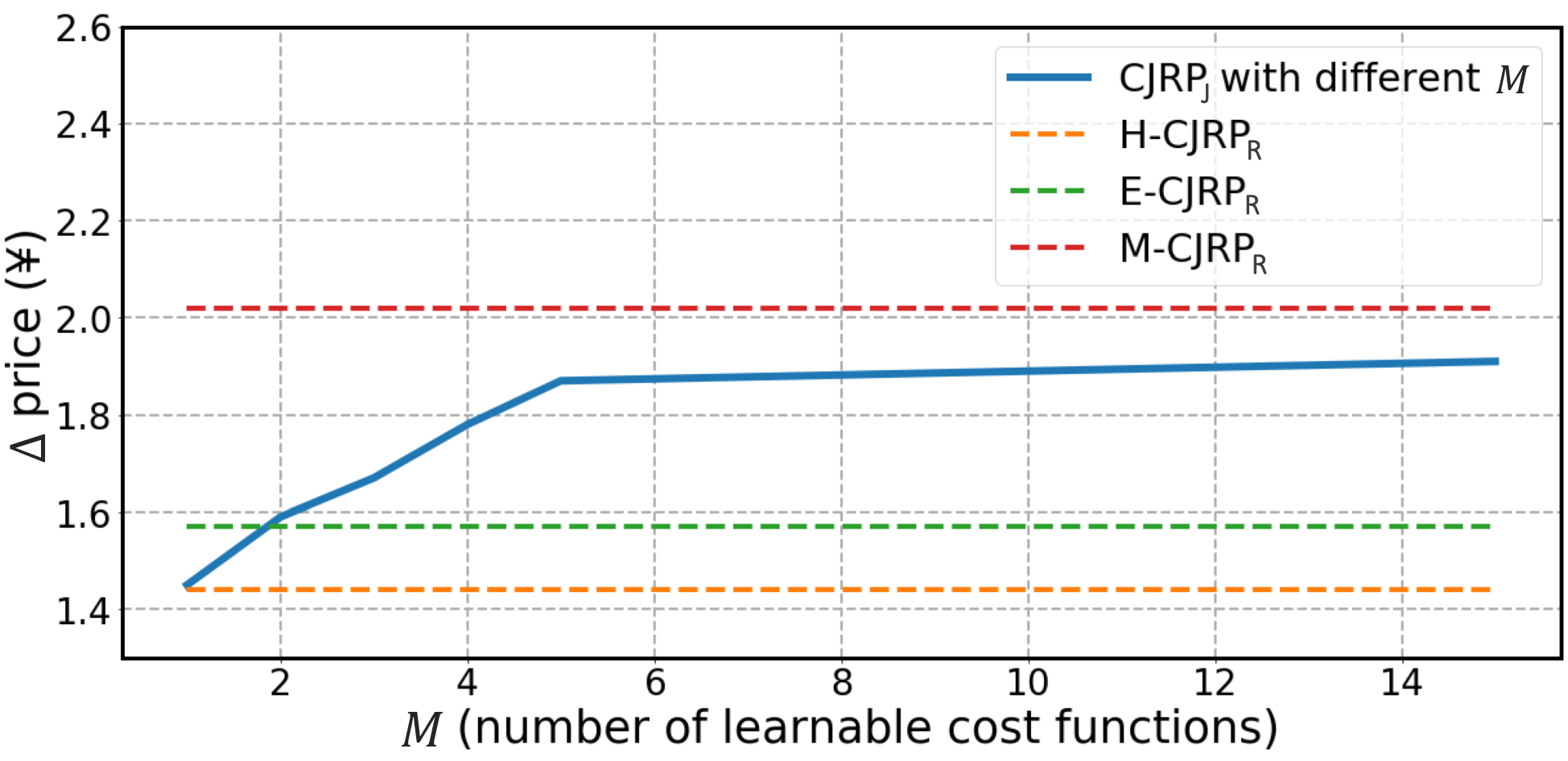} 
	\caption{Effect of the number of learnable cost functions on the economic preference. Y-axis denotes the difference of order price between H-Lambda and the CJRP family.} 
	\label{fig:effect-cf} 
\end{figure}

{\bfseries Effect of the Number of Learnable Cost Functions.} For a preprocessing-based route-planning scheme like CRP, each additional cost function consumes extra computing resources (CPU and memory). To explore the balance between performance and resource cost, we study the effect of the number of learnable cost functions on the economic preference. Figure \ref{fig:effect-cf} shows the results. If we consider the improvement of M-$\text{CJRP}_{\text{R}}$ as the maximum improvement obtainable, then 3 (5) extra cost functions suffice to achieve 80\% (90\%) of the maximum.

\section{CONCLUSION\label{conclusion}}
In this paper, we propose a novel deep architecture for jointly optimizing the cost functions and ranking model in a route-planning system. We run a multi-objective Dijkstra algorithm offline to compute the Pareto-optimal set for each routing request and deem it as the complete candidate set. Exploiting the property of such a set, we design a neural network structure that performs a shortest-path search. This structure, together with a cascaded MLP-based ranking module, emulates the process of route planning in a fully differentiable manner. To further adapt the joint model to the various route preferences, we propose a customizable loss function, constructed with unbiased estimators for route attributes. In addition, we propose a novel batch sampling method that ensures local validity of the constraints. Finally, evaluations on real-world datasets show that our architecture significantly outperforms state-of-the-art methods in both route quality and customizability.


\bibliographystyle{plainnat}
\bibliography{sample-base}

\appendix

\end{document}